# Exploring Aviation Incident Narratives Using Topic Modeling and Clustering Techniques

Aziida Nanyonga
School of Engineering and Technology
University of New South Wales
Canberra, Australia
a.nanyonga@unsw.edu.au

Hassan Wasswa
School of Systems and Computing
University of New South Wales
Canberra, Australia
h.wasswa@unsw.edu.au

Ugur Turhan
School of Science
University of New South Wales
Canberra, Australia
u.turhan@unsw.edu.au

Keith Joiner
School of Engineering and Technology
University of New South Wales
Canberra, Australia
k.joiner@unsw.edu.au

Graham Wild
School of Science
University of New South Wales
Canberra, Australia
g.wild@unsw.edu.au

*Abstract*—Aviation safety is a global concern, requiring detailed investigations into incidents to understand contributing factors comprehensively. This study uses the National Transportation Safety Board (NTSB) dataset. It applies advanced natural language processing (NLP) techniques, including Latent Dirichlet Allocation (LDA), Non-Negative Matrix Factorization (NMF), Latent Semantic Analysis (LSA), Probabilistic Latent Semantic Analysis (pLSA), and K-means clustering. The main objectives are identifying latent themes, exploring semantic relationships, assessing probabilistic connections, and cluster incidents based on shared characteristics. This research contributes to aviation safety by providing insights into incident narratives and demonstrating the versatility of NLP and topic modelling techniques in extracting valuable information from complex datasets. The results, including topics identified from various techniques, provide an understanding of recurring themes. Comparative analysis reveals that LDA performed best with a coherence value of 0.597, pLSA of 0.583, LSA of 0.542, and NMF of 0.437. K-means clustering further reveals commonalities and unique insights into incident narratives. In conclusion, this study uncovers latent patterns and thematic structures within incident narratives, offering a comparative analysis of multiple-topic modelling techniques. Future research avenues include exploring temporal patterns, incorporating additional datasets, and developing predictive models for early identification of safety issues. This research lays the groundwork for enhancing the understanding and improvement of aviation safety by utilising the wealth of information embedded in incident narratives.

## I. INTRODUCTION

Aviation safety is a paramount concern in contemporary society, necessitating investigations into aviation incidents to understand the multifaceted factors contributing to mishaps. The National Transportation Safety Board (NTSB) provides a comprehensive dataset of aviation incident narratives, a treasure trove of information on past incidents. Analyzing this dataset offers an invaluable opportunity to extract nuanced insights, contributing to the continuous enhancement of safety protocols [1]. The vast amount of textual data in aviation incident reports poses a significant challenge for manual analysis due to its volume, making the process time-consuming and impractical. Automated text analysis methods, specifically topic modelling, have emerged as effective solutions to this challenge. These techniques allow the identification of latent thematic structures within the textual data, facilitating the extraction of pertinent information, trends, and patterns [2]–[4].

The NTSB dataset, a repository of incident narratives spanning diverse scenarios and contributing factors, serves as a vital archive for the aviation industry. Each narrative provides a unique account of events, offering a nuanced perspective on the circumstances surrounding incidents. Leveraging advanced Natural Language Processing (NLP) techniques and machine learning methodologies can reveal latent patterns within these narratives, shedding light on recurring themes, hidden correlations, and potential areas for improvement.

This study aims to harness the power of advanced techniques, including Latent Dirichlet Allocation (LDA), Non-negative Matrix Factorization (NMF), Latent Semantic Analysis (LSA), Probabilistic Latent Semantic Analysis (pLSA), and K-means clustering, to delve into the intricate details of aviation incident narratives within the NTSB dataset. The specific objectives include:

1. Utilizing topic modelling techniques to identify latent themes within aviation incident narratives.

2. Applying matrix factorization techniques to decompose the dataset and identify key components.

3. Exploring latent semantic relationships within the narratives using LSA.

4. Assessing the probabilistic relationships between terms and topics through pLSA.

5. Employing K-means clustering to group incidents based on shared characteristics, providing a holistic view of narrative structures.

This research is significant on multiple fronts. Firstly, it contributes to the broader discourse on aviation safety by offering a nuanced understanding of incident narratives. Secondly, it showcases the versatility of advanced NLP and machine learning techniques in extracting valuable information from complex narratives, a methodological contribution to the NTSB dataset. Ultimately, the findings of this study have the potential to inform aviation authorities, empower safety professionals, and inspire further research, fostering a safer aviation environment.

In the subsequent sections, we will delve into the details of the related work (Section II), our methodology, data collection, and the application of advanced techniques, to reveal the untold stories within aviation incident narratives (Section III). The results (Section IV), conclusion, and avenues for future work (Section V) will follow in this exploration.

## II. RELATED WORK

Topic modelling plays a pivotal role in text analysis and NLP, offering a potent method to unveil latent thematic structures within extensive textual data volumes. In recent years, the application of topic modelling techniques and NLP in analyzing aviation incident narratives has gained significant traction. Traditionally, accident investigation and safety analysis relied on expert analysis and statistical methods [2], [5], [6]. However, these methods, relying on manual examination of accident reports, are time-consuming and susceptible to human bias [7].

Advancements in automated text analysis methods, notably in NLP and machine learning, have revolutionized aviation safety research [8]–[10]. Utilizing techniques such as text mining and sentiment analysis, researchers have gained scalability and objectivity while minimizing human bias when scrutinizing reports. Furthermore, within the realm of NLP, topic modelling has emerged as a significant tool in aviation safety research, providing a systematic method for revealing latent thematic structures within textual data. Key algorithms like LDA and NMF have been applied across diverse textual datasets, further enhancing the analytical capabilities in this field [11, 13, 14, 15].

The utilization of topic modelling in the analysis of aviation incident reports presents significant potential, as seen in studies utilizing pLSA and LDA [4]. Effective text preprocessing, such as NLP techniques [17], and methodologies like Latent Semantic Analysis (LSA) [18] and NMF [19], contribute to enhancing the quality of extracted topics.

Noteworthy, research [13], [5], [2], [20], [21] has showcased the potential of computational methods, text mining, and machine learning in accident report analysis, emphasizing efficiency and consistency. Our research builds upon these foundations by conducting a comprehensive comparative analysis of pLSA, LSA, LDA, and NMF, along with k-means clustering, applied to the NTSB dataset. This work contributes to ongoing efforts to enhance aviation safety and risk assessment through advanced text analysis methodologies.

## III. METHODOLOGY

Before In this section, we explain the methodology employed to explore aviation incident narratives within the NTSB dataset. Our approach integrates advanced NLP and ML techniques to unveil latent patterns, hidden correlations, and thematic structures embedded in the narrative. The next subsection gives a brief description of the data sets used in this study.

### A. Data Acquisition

Aviation incident and accident investigation reports used in this study were exclusively sourced from the NTSB dataset spanning years from 2000 to 2020. The dataset comprising a collection of more than 36,000 records in JSON format was obtained from the following source: https://www.ntsb.gov/Pages/AviationQuery.aspx. These reports encompass textual narratives, findings, and recommendations from NTSB investigations, making them an invaluable resource for the study.

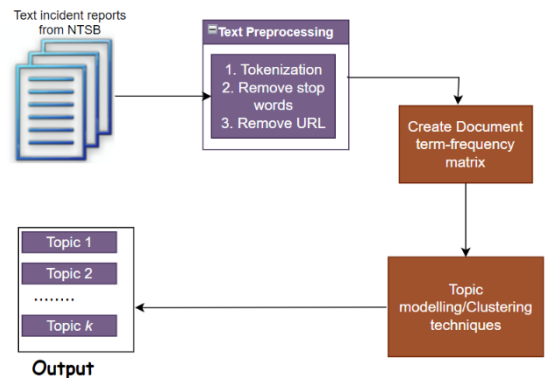

Fig. 1. Methodological framework

### B. Text Processing

Machine learning algorithms cannot inherently comprehend raw textual data. Our text preprocessing pipeline comprises multiple crucial stages aimed at enhancing data quality and optimizing model performance. These stages, illustrated in Fig. 1, encompass tokenization, lowercasing, punctuation removal, stopword elimination, and URL removal. Lowercasing ensures uniformity in text representation, while punctuation removal simplifies text for analysis. Tokenization dissects narratives into individual words, facilitating further analysis. Eliminating stopwords reduces noise, and URL removal prevents web links from interfering with analysis [8].

After undergoing these preprocessing steps, narratives are primed for feature extraction, a pivotal process that transforms textual data into numerical features suitable for machine learning models. For feature extraction, we employed two distinct techniques: Term Frequency-Inverse Document Frequency (TF-IDF) and Word Embeddings (Word2Vec). TF-IDF assesses term importance within narratives, capturing semantic meaning. Word2Vec represents words as dense vectors, enabling models to comprehend semantic relationships. Supplementary preprocessing steps included HTML tag removal, non-alphanumeric character elimination, and exclusion of irrelevant elements. Finally, lemmatization standardizes words to their base forms, enhancing interpretability in topic modelling.

### C. Topic Modeling Techniques

*1) Latent Dirichlet Allocation (LDA):* LDA is employed to identify latent topics within the narratives. The algorithm models each document as a distribution of topics and each topic as a distribution of words. This facilitates the extraction of themes inherent in the narratives. Fig. 2 illustrates the operational mode of an LDA model. The model utilizes a three-phase stochastic approach to allocate topics to clusters of words within each document. Initially, in phase one, topics are sampled from a Dirichlet distribution for each document [21]. Subsequently, during phase two, each word in the document is assigned a topic from the sampled topics obtained in phase one. Finally, in phase three, each word assigned to a topic in phase two is sampled from a multinomial distribution over words associated with that topic. In this model, the matrix φ represents the topic distributions, with a multinomial distribution over N-word items for each of T topics drawn independently from a symmetric Dirichlet (β) prior. θ represents the matrix containing document-specific mixture weights for a set of T topics, with each weight drawn independently from a symmetric Dirichlet (β) prior distribution. Within this framework, z identifies the topic attributed to generating a particular word, drawn from the θ distribution specific to the document, while w denotes the word itself, drawn from the topic distribution φ corresponding to z. Nd represents the total number of words in the document, and D signifies the size of the document collection. The estimation of φ and θ yields insights into the topics present in the collection and their respective weights within each document.

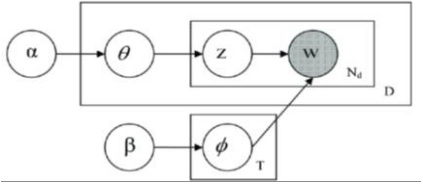

Fig. 2. Graphical representation of the LDA model [21]

*2) Non-Negative Matrix Factorization (NMF):* NMF is utilized for matrix factorization, decomposing the dataset into non-negative matrices representing key components. This technique aids in revealing underlying structures within the narratives. In our analysis, we also employed Non-negative Matrix Factorization (NMF) as an alternative approach for dimensionality reduction in topic modelling. NMF decomposes the Document-Term Frequency Matrix into two lower-dimensional matrices: one representing topics and the other representing term distributions [17]. What distinguishes NMF is its inherent interpretability, which proves valuable in extracting meaningful insights from aviation accident reports. The dataset is represented as a matrix V with dimensions w×d, where w denotes words in each document and d represents documents. Fig. 3 illustrates a simplified depiction of how NMF decomposes V into its constituent parts, W and H, where W is a matrix of dimensions $w \times t$, H is a matrix of dimensions $t \times d$, and t represents the distinct topics in V.

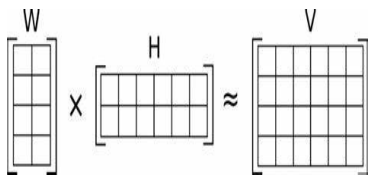

Fig. 3. Non-negative matrix Factorization diagram

*D. Semantic Analysis*

*1) Latent Semantic Analysis (LSA):* LSA is applied to capture latent semantic relationships within the narratives. By representing words and documents in a high-dimensional semantic space, it unveils nuanced connections between terms. LSA, as a method of dimensionality reduction, converts the document-term matrix into a space with fewer dimensions. It reveals the inherent structure of text data by discerning connections between terms and documents [22]. In our study, LSA was employed to extract topics from aviation safety reports, thus offering insights into the concealed themes present in the corpus.

*2) Probabilistic Latent Semantic Analysis (pLSA):* pLSA is employed to model the probabilistic relationships between terms and topics. This probabilistic approach allows for a more nuanced understanding of the uncertainty inherent in incident narratives. pLSA takes a probabilistic approach, contrasting with Singular Value Decomposition (SVD) in tackling the topic modelling problem [23]. It constructs a probabilistic model featuring latent topics to generate the observed data in the document-term matrix. Specifically, it aims to establish a model $P(D, W)$ that assigns probabilities to each entry in the document-term matrix for any given document D and word W. Adhering to the foundational assumptions of topic models, which propose that each document is a mixture of various topics and each topic comprises a set of words, pLSA introduces a probabilistic interpretation to these principles:

1. For document d, pLSA assigns topic z to that document with the likelihood denoted as $P(z|d)$.
2. When considering a topic z, pLSA models the probability of drawing a word w from that topic as $P(w|z)$.

The model is represented as

$$P(D, W) = P(D) \sum_z P(z|d) P(w|z)$$

where P(D), $P(Z|D)$, and $P(W|Z)$ are parameters. P(D) can be directly computed from the corpus data, while $P(Z|D)$ and $P(W|Z)$ are represented as multinomial distributions and can be trained using the expectation-maximization algorithm (EM) [24]. In simple terms, EM is a technique used to determine the most probable parameter estimates for a model relying on unobserved, latent variables (in this context, the topics).

*E. K-means Clustering*

K-means clustering is a widely used unsupervised machine learning technique that aims to partition a dataset into K distinct, non-overlapping subsets or clusters. Each data point is assigned to the cluster with the nearest mean, and the algorithm iteratively refines the cluster assignments to minimize the within-cluster variance [25].

The algorithmic overview of the K-means is as follows:

- Randomly select K data points as the initial centroids of the clusters.
- Assign each data point to the cluster whose centroid is the nearest, typically using Euclidean distance.
- Recalculate the centroids based on the meaning of the data points within each cluster.
- Iterate the assignment and centroid update steps until convergence, which occurs when the assignments no longer change significantly and the result is K clusters, each characterized by its centroid.

*F. Evaluation Metrics*

To assess the chosen techniques' effectiveness, coherence scores and interpretability metrics are employed. Coherence scores measure semantic similarity between high-probability words in topics, while interpretability metrics evaluate the human interpretability of generated topics.

## G. Implementation

The execution of our methodology is carried out using the Python programming language with Jupyter Notebook serving as the integrated development environment (IDE). We leverage the capabilities of prominent libraries, including NLTK, Gensim, and Scikit-Learn. Each technique is applied iteratively, with parameters fine-tuned to optimize performance. The selection of these libraries is grounded in their robust functionalities in natural language processing (NLP) and machine learning (ML), ensuring a comprehensive and efficient implementation of our techniques. This approach not only fosters transparency but also encourages reproducibility and extension of our analyses by future researchers.

## IV. RESULTS AND DISCUSSION

In this section, we present the outcomes of our analysis, including the identified topics from various techniques, NMF, PLSA, LSA, and LDA.

### A. Topic Extraction and Coherence Evaluation.

We employed the Coherence Value (C_v) as an evaluation metric to assess the quality of topics generated by the models. The C_v measures the semantic coherence of topics, with higher values indicating more coherent topics. In our study, LDA yielded a C_v coherence score of 0.597, while outperforming other models as shown in Table 1.

TABLE I. SHOWS THE COHERENCE SCORE OF ALL MODELS.

| Technique | Coherence Value |
| --- | --- |
| Non-Negative Matrix Factorization (NMF) | 0.437 |
| Latent Dirichlet Allocation (LDA) | 0.597 |
| Latent Semantic Analysis (LSA) | 0.542 |
| Probabilistic Latent Semantic Analysis (PLSA) | 0.583 |

### B. NMF and LDA model's performance

The topic distribution from the document, as illustrated in Fig. 4, showcases the contrasting approaches of Latent Dirichlet Allocation (LDA) and Non-Negative Matrix Factorization (NMF) models. In Fig.4, the LDA model offers a comprehensive view of topic distribution across all topics providing a holistic perspective on the document's content. Conversely, the NMF model selectively emphasizes topic distribution primarily focusing on topics 4, 6, and 8.

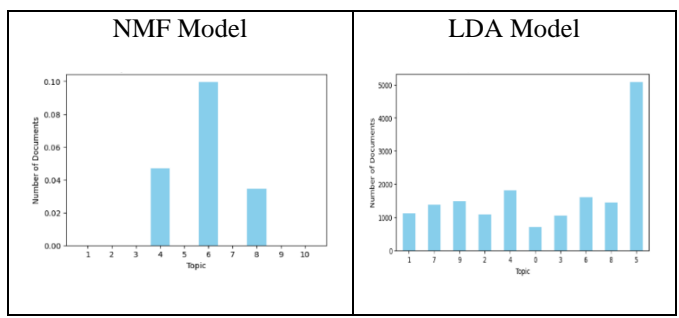

Fig. 4. Shows the topic distribution of the words from the document on LDA and NMF models

### C. Top Words for Each Topic and Word Clouds

We present an overview of the identified topics, their associated words, the distribution of topics across the dataset, and a thematic word cloud visualization. Examining Fig. 5, which depicts the top words for each topic chosen by the pLSA model, it becomes evident that certain topics exhibit higher word scores than others. Notably, the words associated with topic three seen in Fig. 4 stand out with elevated word scores, indicating their prominence within the document. This nuanced insight into word importance contributes to a more detailed understanding of the topics extracted by the LDA model.

### D. pLSA and LSA models performance

LSA leverages mathematical techniques to uncover underlying themes within aviation incident narratives. We present an overview of the identified topics, their associated words, and the proportion of variance explained, as well as a thematic word cloud visual representation of these topics. Also, Fig. 6 shows the term weights of words selected by LSA on the first four topics. This visualization helps to determine the optimal number of topics to capture the underlying thematic structure of the aviation incident narratives.

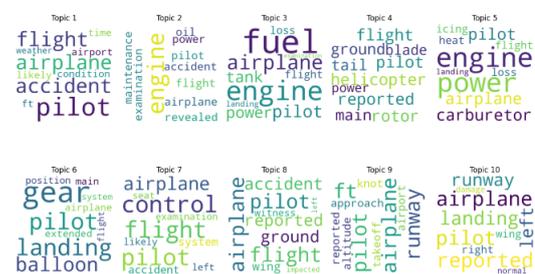

Fig. 5. 10 words that were chosen by the pLSA model in each Topic

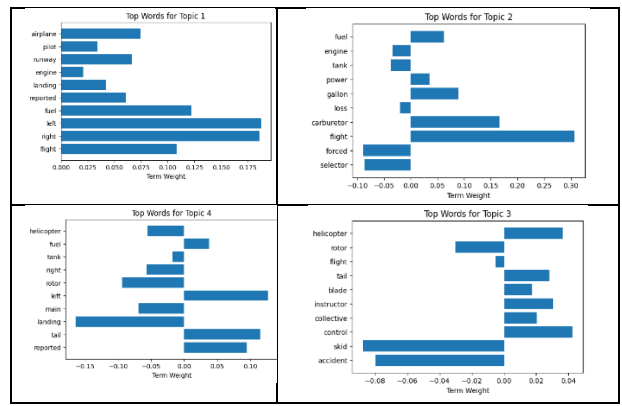

Fig. 6. Shows the term weights of the words that were selected by LSA on the first four topics.

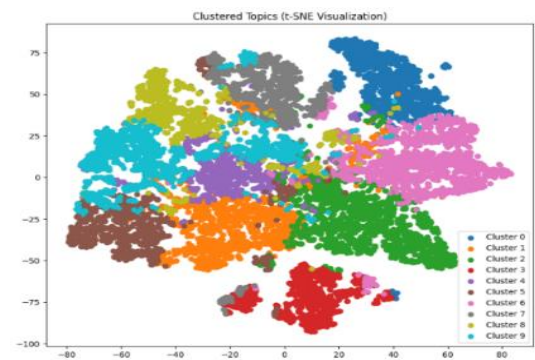

Fig. 7. Clustered Topics (t-SNE Visualization).

TABLE II. SHOWS THE TOPIC INTERPRETATION FOR ALL MODELS.

| No. Topics | NMF chosen Topics | Lda chosen topics | LSA chosen Topics | pLSA chosen topics |
|---|---|---|---|---|
| Topic 1 | Mechanical Failures | Maintenance and Inspection | Pilot-Centric Elements | Cabin Fumes and Engineering |
| Topic 2 | Fuel System Issues | Helicopter Operations | Fuel System Dynamics | ATC and Aircraft Separation |
| Topic 3 | Engine and Power Loss | Fuel System and Pilot Actions | Flight and Helicopter Dynamics | Engine Failure in Cruise |
| Topic 4 | Helicopter Operation | Landing Gear and Aircraft Control | Landing Gear and Helicopter Rotor | Bird Strikes |
| Topic 5 | Landing Gear Incidents | Engine and Power Loss | Engine and Fuel Interaction | Safety Officer Reports |
| Topic 6 | Flight Conditions | Accident Conditions | Runway and Flight Dynamics | Runway Incidents |
| Topic 7 | Student Pilot Incidents | Aviation Accident Dynamics | Instructor-Led Flight | Rejected Takeoff |
| Topic 8 | Wind and Weather Impact | Weather and Airport Conditions | Power Issues | Minor Damage Incidents |
| Topic 9 | Rudder and Brake Issues | Takeoff and Runway Conditions | Airport and Runway Dynamics | GPWS Alerts |
| Topic 10 | Runway Incidents | Passenger and Glider Dynamics | Runway and Aircraft Dynamics | Routine Inspection Findings |

### E. Comparative Analysis and K-means Clustering

*1) Common Themes Across Models:* Despite the differences in the algorithms, certain themes consistently emerge across multiple models as shown in Table 2. For instance, topics related to "Mechanical Failures," "Fuel System Issues," and "Engine and Power Loss" are recurrent, suggesting their prevalence in aviation incidents. This consistency reinforces the significance of these factors in contributing to safety concerns. In interpreting these results, it is essential to note the nuanced patterns and recurring themes across different models. While NMF, LSA, pLSA, and LDA offer distinct perspectives on the dataset, the convergence of certain keywords across topics suggests common threads in aviation incident narratives. The identified topics provide a foundation for further exploration and analysis, contributing valuable insights to the broader discourse on aviation safety.

*2) Unique Insights from Each Model:* While common themes provide a baseline understanding, each modelling technique offers unique insights. NMF emphasizes specific aspects such as "Helicopter Operation" and "Flight Conditions," shedding light on scenarios that might be overlooked by other techniques. LSA captures nuanced relationships, revealing connections between "Wind and Weather Impact" and "Rudder and Brake Issues.". Table 3 depicts the top 10 words that were chosen by all models among the 10 topics.

### F. K-means Clustering of Incident Narratives

In addition to topic modelling, K-means clustering is employed to categorize incidents based on shared characteristics, revealing underlying patterns and providing a complementary perspective to topic modelling. This approach enhances the interpretability of incident narratives by grouping them into distinct clusters, each defined by prevalent characteristics and keywords. These clusters, highlighted through K-means clustering, offer a structured view of narrative patterns, contributing to a more nuanced understanding of incident types and facilitating targeted interventions for safety improvement.

Clusters are formed by grouping incidents with similar attributes enabling a more focused analysis of shared characteristics. Each cluster encapsulates a set of incidents with commonalities, providing a clearer delineation of narrative patterns. For instance, a cluster may Centre around "Runway Incidents" or "Landing Gear and Helicopter Rotor," highlighting the thematic coherence within each group. Fig. 12 visually represents the clustered topics using t-SNE (t-distributed Stochastic Neighbor Embedding) visualization. The figure illustrates ten distinctive clusters. The colour blue (Cluster 0), orange (Cluster 1), and green (Cluster 2) respectively provide a visual representation of the incident narratives grouped by K-means clustering. This visualization aids in identifying patterns and relationships between incidents within and across clusters, offering a valuable perspective for safety analysis.

### V. CONCLUSION

This research employs advanced topic modelling and clustering techniques to delve into aviation incident narratives within the NTSB dataset. Through a comparative analysis of NMF, LSA, pLSA, and LDA, the study uncovers both common and unique insights, providing an understanding of recurring themes in aviation incidents. K-means clustering further enhances this comprehension by organizing incidents into meaningful groups, offering a nuanced perspective on narrative structures.

The study significantly contributes to aviation safety research by unveiling latent patterns and thematic structures within incident narratives. The exploration of incident clusters enhances the roughness of safety analysis, providing valuable insights into the complexities of aviation safety.

As avenues for future research, temporal patterns in incident narratives could be explored to identify evolving safety concerns. Furthermore, the incorporation of additional datasets could validate and expand the findings, offering a broader perspective on aviation safety. Additionally, the development of predictive models for the early identification of potential safety issues presents a promising direction for advancing the field. Therefore, this study acts as a foundational step for further advancements in understanding and improving aviation safety, leveraging the wealth of information embedded in incident narratives.

TABLE III. TOP 5 WORDS FOR EACH OF THE 10 TOPICS THAT WERE CHOSEN BY ALL MODELS

| Models | Topic 1 | Topic 2 | Topic 3 | Topic 4 | Topic 5 | Topic 6 | Topic 7 | Topic 8 | Topic 9 | Topic 10 |
|---|---|---|---|---|---|---|---|---|---|---|
| NMF | Right | Fuel | Engine | Helicopter | Gear | Light | Student | Wind | Reported | Runway |
| | Left | Tank | Power | Rotor | Landing | Accident | Instructor | Knot | Mechanical | Approach |
| | Brake | Gallon | Carburetor | Tail | Nose | Condition | Flight | Gust | Malfunction | Takeoff |
| | Rudder | Engine | Loss | Blade | Collapsed | Witness | Control | Accident | Operation | Stated |
| | Wing | Selector | Oil | Collective | Approach | Instrument | Landing | Weather | Failure | Final |
| LDA | Maintenance | Helicopter | Fuel | Landing | Engine | Airplane | Airplane | Pilot | Airplane | Pilot |
| | Revealed | Flight | Tank | Gear | landing | Pilot | Pilot | Flight | Runway | Line |
| | Inspection | Instructor | Engine | Airplane | Power | Reported | Accident | Airplane | Pilot | Aircraft |
| | Manufacturer | Pilot | Pilot | Pilot | Pilot | Runway | Flight | Accident | Takeoff | Glider |
| | Examination | Rotor | Airplane | Brake | Airplane | Landing | Engine | Condition | Approach | passenger |
| LSA | Airplane | Fuel | Helicopter | Helicopter | Gear | Gear | Pilot | Left | Runway | Runway |
| | Pilot | Engine | Rotor | Fuel | Landing | Flight | Flight | Helicopter | Flight | Accident |
| | Runway | Tank | Flight | Tank | Engine | Landing | Helicopter | Right | Instructor | Flight |
| | Engine | Power | Tail | Rotor | Power | Helicopter | Left | Landing | Airplane | Helicopter |
| | Landing | Gallon | Instructor | Landing | Carburetor | Reported | Instructor | Gear | Reported | Pilot |
| PLSA | Pilot | Engine | Fuel | Helicopter | Engine | Gear | Control | Airplane | Airplane | Airplane |
| | Airplane | Airplane | Engine | Pilot | Power | Landing | Pilot | Pilot | Pilot | Pilot |
| | Flight | Pilot | Airplane | Rotor | Carburetor | Pilot | Flight | Flight | Runway | Reported |
| | Accident | Revealed | Pilot | Flight | Pilot | Balloon | Airplane | Accident | Approach | Landing |
| | Airport | Examine | Power | Reported | Airplane | Airplane | System | Reported | Knot | Runway |